# Mitosis Detection in Intestinal Crypt Images with Hough Forest and Conditional Random Fields

Gerda Bortsova[1,2], Michael Sterr[3], Lichao Wang[1,2], Fausto Milletari[2], Nassir Navab[2,4], Anika Böttcher[3], Heiko Lickert[3], Fabian Theis[1,5,*] and Tingying Peng[1,2,5,*]

[1]Institute of Computational Biology, Helmholtz Zentrum München, Germany
[2]Computer Aided Medical Procedures, Technische Universität München, Germany
[3]Institute of Diabetes and Regeneration Research, Helmholtz Zentrum München, Germany
[4]Computer Aided Medical Procedures, Johns Hopkins University, USA
[5]Chair of Mathematical Modelling of Bioloigcal Systems, Technische Universität München, Germany
[*]Correspondence: fabian.theis@helmholtz-muenchen.de, tingying.peng@tum.de

**Abstract.** Intestinal enteroendocrine cells secrete hormones that are vital for the regulation of glucose metabolism but their differentiation from intestinal stem cells is not fully understood. Asymmetric stem cell divisions have been linked to intestinal stem cell homeostasis and secretory fate commitment. We monitored cell divisions using 4D live cell imaging of cultured intestinal crypts to characterize division modes by means of measurable features such as orientation or shape. A statistical analysis of these measurements requires annotation of mitosis events, which is currently a tedious and time-consuming task that has to be performed manually. To assist data processing, we developed a learning based method to automatically detect mitosis events. The method contains a dual-phase framework for joint detection of dividing cells (mothers) and their progeny (daughters). In the first phase we detect mother and daughters independently using Hough Forest whilst in the second phase we associate mother and daughters by modelling their joint probability as Conditional Random Field (CRF). The method has been evaluated on 32 movies and has achieved an AUC of 72%, which can be used in conjunction with manual correction and dramatically speed up the processing pipeline.

## 1  Introduction

The intestinal epithelium is the most vigorously renewing adult tissue in mammals. The intestinal stem cells (ISCs) located at the bottom of the crypts fuel this process [1]. Under normal conditions, ISCs are maintained by symmetric self-renewal and undergo neutral competition to contact their supporting niche cells. Upon loss of short range niche signals, ISCs can be under differentiation [2]. Nevertheless, asymmetric modes of ISC division, contributing to ISC homeostasis and secretory progenitor commitment, are under debate [3]. The differentiation of the secretory lineage is of particular interest, since it comprises the enteroendocrine cells, which secrete various hormones, involved in energy and glucose homeostasis [4]. In recent years, the role of enteroendocrine cells in development and treatment of diabetes is increasingly recognized, but the exact mechanism of their differentiation remains unclear.

In order to investigate the mechanisms underlying enteroendocrine differentiations, we monitor dynamic cell division and differentiation of murine intestinal crypt using confocal microscope. Based on these live cell movies, we can correlate cell division modes to specific image features. For example, symmetrically divided daughter cells have a very similar cell shape and appearance, whilst asymmetrically divided daughter cells, by contrast, tend to have different sizes, shapes and appearance. Particularly, in a typical asymmetric division case, only one daughter cell touches the crypt outer membrane (as shown in Fig. 1).

So far, these live cell movies are inspected manually, which is laborious and time-consuming. Hence we aim to develop an automatic processing pipeline to accelerate the analysis, in which the key component is an automatic detection of cell division (mitosis) for these movies.

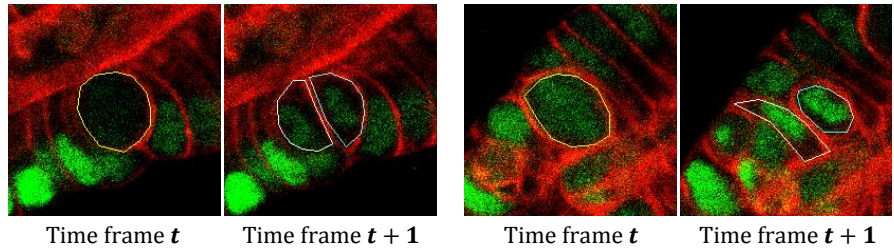

| Time frame $t$ | Time frame $t+1$ | Time frame $t$ | Time frame $t+1$ |

**Fig. 1.** Exemplary symmetric (left image pair) and asymmetric cell divisions (right pair). Cell membranes are visualized in red channel and nuclei are shown in green channel. Mothers are outlined with yellow contour, daughters with white and blue. The symmetrically divided daughters have similar shape and appearance, whilst the asymmetrically divided daughters have different appearance.

## 2    Problem definition

Unlike "mitosis detection" of histology images where mitotic cells are identified based on one single static image [5], our goal is to detect "mitosis" as a dynamic process, i.e., to identify both mother cell right before the division and daughter cell pair right after the division (Fig. 1).

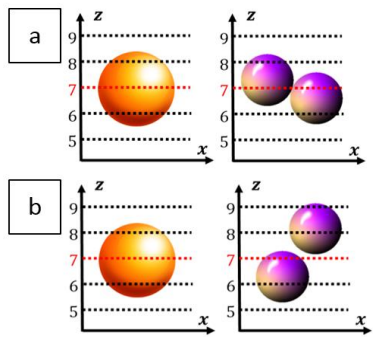

**Fig. 2.** Daughters may stay on the same z-plane as their mother (a) or move to other planes (b).

The first challenge that we are facing is the time resolution in our movie is rather low (15 minutes between frames; higher time resolution often results in cell death). With this temporal resolution, one cannot see all the stages of the division (e.g. elongation of the cell) but rather an instant mother to daughter cell splitting between two consecutive frames (also shown in Fig. 1). Compared to mother cells which usually have a characteristic round shape, daughter cells are much less distinguishable from other normal cells and also have larger variability of their shape and appearance, which makes the identification of daughter cells difficult. Besides, the time gap also results in a significant frame-to-frame cell movement as well as variations in their shape and appearance.

Another significant challenge is the poor z-axis resolution (≈25 times smaller than in x and y) of our dataset which makes 3D cell detection almost impossible as daughter cells can be viewed well mostly at one z-plane. Cell divisions, however, do happen in 3D, so the daughters may stay on the same z-plane with their mother, and may migrate to other z-planes (Fig. 2). In the latter case it is difficult to confidently track daughter cells even for a human expert. Hence, we don't consider this case in our detection, and focus only on the case when the mother and both daughters are visible in the same z-plane, which makes our detection essentially 2D. Since our ultimate goal is to quantify the ratio of symmetric and asymmetric cell divisions in crypt, ignoring out-of-plane cell division is not a problem, as the ratio should stay the same when large amount of movies are analysed.

Rapid frame-to-frame cell movement and low z-resolution create a situation where roughly in half of the cases a mother cell is identified but at least one of its two daughters cannot be confidently traced. Such a case is not considered as a complete "mitosis event" for us (as we need both mother and daughter pair).

## 3  Methods

In this paper, we propose a dual-phase scheme for mitosis detection (shown in Fig. 3). The input to the algorithm is two consecutive time frames belonging to the same z-plane. The goal of the first phase is to obtain two probability maps: the probability of location of a mother on the first frame and a daughter pair on the following one. The method to obtain these maps is described in section 3.1. In the second phase we use a joint probability distribution modelled by Conditional Random Field (CRF) in order to detect mitosis events by matching candidate mother and daughter pair. This is explained in details in section 3.2.

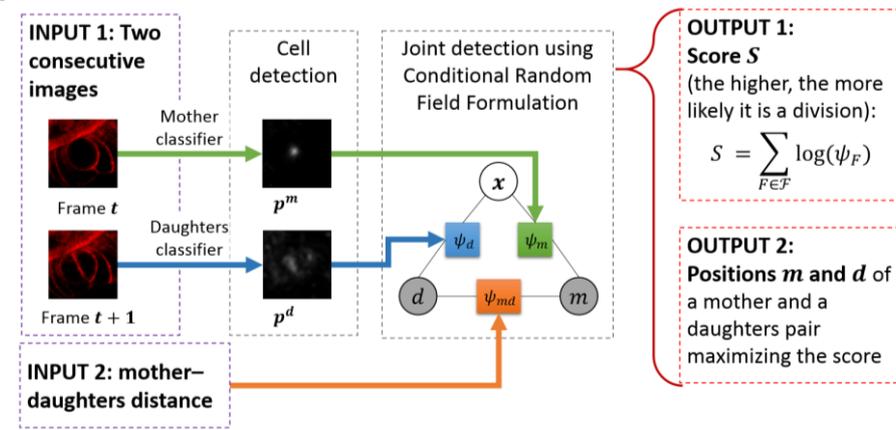

**Fig. 3.** The dual-phase mitosis detection scheme.

### 3.1 Cell detection with Hough forest

Cell detection is an important and classic research topic in automatic bioimage processing. Among proposed learning-based methods, one approach treats cell detection as a classification problem and trains a classifier to identify cell centroids (e.g. [6]). The classifiers are applied either densely on images or on candidate regions found by e.g. blob detectors or other classifiers [7]. This approach could be problematic in case of a small dataset, i.e. number of positive samples would be also small and hence lead to a high overfitting risk. Another approach is to formulate the problem as regression problem and learn a distance map to a closest cell's centre [8]. Regression solves the problem of having several detections per cell which classic classification might have and hence achieves better performance. A similar method that has this advantage is Hough forest [9], which also learn displacement vectors to object's centre. Compared to classic classification forest, Hough forest does not only split based on information gain, but also splits to increase the uniformity of the displacement vectors of the positive points (negative points do not participate in the voting). Therefore, Hough forest implicitly enforces shape constraints on objects and conveniently provides both segmentation of an object and its centre. Furthermore, unlike regression forest, Hough forest is easily extended to multiclass (which is the case of our problem as we need to detect two kinds of cells: mother and daughter). Although being previously used for many computer vision problems, Hough forest, to our best knowledge, hasn't been used for cell detection so far.

The class distribution of our learning problem is highly imbalanced as the foreground points (positive) constitute less than 0.5% of all points. Instead of creating a balanced training dataset which the variability of negative class is under-represented, we weight the class frequencies $n^l$ by calculating a probability of assigning a sample in leaf $l$ to class $c$ with inverse prior class probabilities $p^{prior}$:

$$p_c^l = \frac{n_c^l / p_c^{prior}}{\sum_{i \in \mathcal{C}} n_i^l / p_i^{prior}}$$

In classification splits, a splitting function minimizing Shannon entropy is chosen:

$$\mathbb{H}(p^l) = \sum_{c \in \mathcal{C}} p_c^l \log p_c^l$$

In vote uniformity splits, a sum of Euclidean distances of the votes $\mathcal{V}(l)$ from the mean vote $\mu_{\mathcal{V}(l)}$ is minimized:

$$VS(l) = \sum_{v_i \in \mathcal{V}(l)} \| v_i - \mu_{\mathcal{V}(l)} \|$$

### 3.2 Joint detection of mother and daughters using CRF

Conditional Random Fields (CRF) were successfully used for joint object detection problems such as human pose estimation [10] or anatomical landmarks detection [11]. This is because that individual object detection might produce many false positives or generate impossible object configuration, enforcement of constraints on geometrical relationships between objects by CRF usually improves the performance. In our prob-lem, daughter cell detection is inherently challenging, as they are not easy to be distinguished from normal cells and have large shape variability. Hence we use the geometrical constraints on daughter cells (our observation is that daughters are usually not too far away from their mother) to reduce their false detection rate.

Our CRF has two random variables: one denoting a position of the mother cell centre (M) and the other denoting a position of the daughter pair centre (D) (see Fig. 3). A feature vector X denotes our Hough maps from individual cell detection. Hence, the joint probability of a location of a mother at a candidate point $m$ and a daughter pair at $d$ follows Gibbs distribution:

$$p(D = d, M = m | X = x) = \frac{1}{Z} \psi_m(m|x) \psi_d(d|x) \psi_{md}(m, d|x) \quad (1)$$

The unary potentials depend on Hough maps from mother and daughter detection:

$$\psi_m(m|x) = \exp(w_m h_m(m)), \psi_d(d|x) = \exp(w_d h_d(d)) \quad (2)$$

The binary potential depends on the Euclidean distance between $m$ and $d$:

$$\psi_{md}(m, d|x) = \exp(w_d \mathcal{N}(\|m - d\| | \mu, \sigma)) \quad (3)$$

The distance is converted to a probability by modeling distance as a normal distribution. Mean and standard deviation of mother-daughter distances are calculated from the training set.

We use the mitosis score $S$ to select good mitosis candidates. It is based on logarithm of the CRF probability distribution:

$$\log(p(m, d|x)) = -\log Z + w_m h_m(m|x) + w_d h_d(d|x) + w_{md} p_{dist}(m, d|x)$$
$$S = w_m h_m(m|x) + w_d h_d(d|x) + w_{md} p_{dist} \quad (4)$$

The normalizing constant $Z$ is omitted due to the fact that it is the same over a neighborhood in the images. Hence the mitosis score turns out to be a simple weighted sum of Hough scores and the probability of distance. In some works in joint object detection the weights of potentials are empirically chosen [12]. However, weighting can greatly affect the results and tuning it manually is usually difficult. A better approach is to learn the weights [11]. In our work, the weights of the CRF potentials are learned using logistic regression.

In order to obtain the maximum a posterior (MAP) estimation of the cell positions in a neighbourhood of an image, we try all possible combinations of mother centre position candidates and daughters centre position candidates by thresholding respective Hough maps, and take one with a highest score. Note that the positions with the highest score are also the ones with the highest probability.

## 4 Results

### 4.1 Experiment and dataset description

Crypts, isolated from the small intestine of Foxa2-Venus fusion; mT/mG mice, were cultured in matrigel for 4 days prior to live cell imaging. Crypts were then imaged using a HC PL APO 20x/0.75 IMM CORR CS2 objective (Leica) on a confocal laser scanning microscope (Sp5, Leica) in bidirectional mode (400 Hz), with 2x line averaging per channel at a resolution of 1024x1024 pixels in x-y direction and a z-step size of 3.99 µm. The resulting voxel size is 0.15 µm x 0.15 µm x 3.99 µm and the time resolution is approximately 15 min. Venus and tdTomato fluorescence were detected simultaneously. We annotated 505 images from 32 different movies, which consists of 424 mother and 233 pairs of daughter cells (233 mitosis events).

### 4.2 Evaluation of cell detection

Firstly, we would like to evaluate our Hough forest based cell detection method, by comparing our Hough forest detection (splitting based on both information gain and voting uniformity, denoted as HF), classification forest detection (splitting based on information gain only) with a subsequent centroid Hough voting (CF+HV), and classification forest detection without any centroid voting (CF). A detected cell centre is considered a true positive (TP) if it is located inside a contour of a segmented cell. In case if there are multiple detections within one contour, one is considered as a TP and all the others are counted as false positives (FPs). Any detection that does not fall into a contour of a cell of a given class (mother or daughter) is considered to be a FP. Contours without any matching detections are false negatives (FNs).

Both the first and the second methods generate a Hough voting map of the objective centres, i.e. the centre of a mother cell and the centre of a daughter cell pair (although in the second experiment we do not attempt to increase the voting uniformity in the leaves, but we still store the displacement vectors associated with positive sampling points and hence create Hough maps). Cell detections are obtained by non-maximum suppression of the Hough maps followed by thresholding. The purpose of the comparison of these two approaches is to understand how much the explicit vote uniformity optimization is contributing to the performance. In the third experiment we obtained detections by applying connected components algorithm on segmentations produced by the classification forest and extracting components' centres. This experiment is done to evaluate the effectiveness of shape enforcement by Hough voting to filter out the false positive points (the predicted positive points that do not accumulate high amount of votes are very likely to be FPs). All three compared methods are computed with 8 trees with maximum height of 19. At each split we evaluate 500 features and 50 thresholds for each feature, minimum number of samples per leaf is set to 10. We use Haar-like features that are computed on the fly with integral images.

All three methods are evaluated by the Precision-Recall curve and the area under the curve (AUC) for both mother detection (Fig. 4, left) and daughter detection (Fig. 4, right). Each point on the curve is the average of 5-fold cross-validation. The training and testing images are sampled on different movies. The curves were obtained by varying threshold on Hough maps (first and second experiments) and on connected components (third experiment).

As shown in Fig. 4, the performance of daughter detection is not as good as that of mother. This is expected as daughter detection is much more challenging due to a number of factors, such as less distinguishing appearance, larger shape variability as well as a much smaller training set (only half of mothers).

As for the comparison of different methods, classic classification forest approach gives substantially worse results (mother AUC: 32.5%, daughter AUC: 11.4%) as compared to the other two methods. It is also expected, as classification alone is not robust against touching cells and can also result in fragmented foreground where either only parts of a cell are presented or they are not in a corrected spatial ordering. Hough forest resolves these issues by implicitly controlling cell shapes using centroid voting. Particularly, optimisation of vote uniformity in Hough forest leads to a further improvement compared to only casting non-optimised votes of training samples in classification forest plus Hough voting (mother AUC: 84.6% vs. 82.1%, daughter AUC: 53.9% vs. 38.4%).

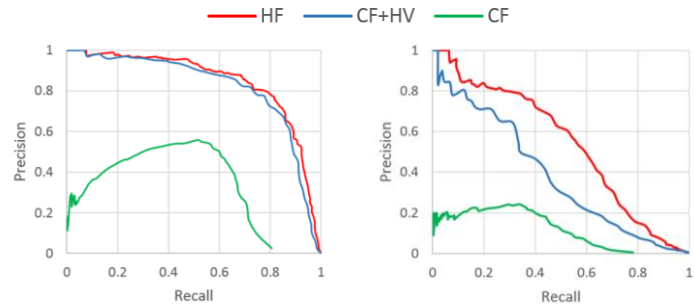

**Fig. 4.** Precision-Recall curves of mother cell detection (left) and daughter pair detection (right) with three compared methods.

### 4.3 Evaluation of mitosis detection

The performance of the mitosis detector is also evaluated using Precision-Recall curve. Based on a similar principle as the cell detector, a TP mitosis event is counted only when the detected mother and daughter pair positions are both inside their corresponding ground truth segmented contours. In case of multiple detections per mitosis event, one is considered as a TP and the rest are FPs. Other detection cases are FPs, and unmatched contours are FNs.

As explained in section 3.2, our joint detection takes into an account three components: mother and daughter pair detection, both in the form of Hough voting map values at candidate positions, as well as a Euclidean distance between these positions converted into a probability by assuming Gaussian distribution. We evaluate the contribution of each component using weights learned from the training data (Eq. 4). Fig. 5 plots the Precision-Recall curve of the mitosis detection obtained by 32-fold cross-validation (leave one movie out). It illustrates that the full model with all three components achieves the best accuracy (AUC: 72.4%). By contrast, performance of reduced models where only two components are considered is lower (mother+daughter: 66.5%, daughter+distance: 61.3%, mother+distance: 30.8%).

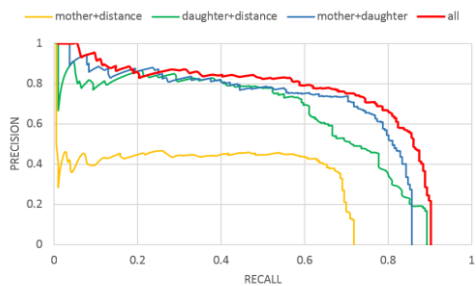

**Fig. 5.** Mitosis detection with CRF.

## 5      Discussion

At present, our proposed dual-phase detection framework achieves a mitosis detection accuracy of 72.4% (AUC), which is very exciting for this challenging problem. The detection accuracy is expected be further improved by providing more annotations, particularly in terms of daughter cells, which have a larger variability in appearance. With current performance, we can already use our automatic mitosis detection algorithm as a pre-processing step and false detections can be manually corrected. Compared to pure manual annotation in which every frame has to be inspected, this pre-processing can dramatically accelerate the analysis process by narrowing manual assessment down to a very small subset of images. For example, if we take 80% recall (detects 80% true mitosis events, we would detect 230 events with 70% precision, that is, around 460 frames (2 frames per event) need to be manually inspected, which is less than 5% of original workload (32 movies, each contains three central z-planes where mitosis events are concentrated and 60-140 frames per plane).

Our present detection algorithm provides us only with the centre of a daughter pair and the corresponding segmentation from Hough forest is not very accurate, as daughters are not separated. This is not sufficient to automatically extract features from daughter cells (such as their size, shape and orientations). So the next step is to develop new processing algorithms to recover more accurate daughter cell segmentation.